# Study and Observation of the Variation of Accuracies of KNN, SVM, LMNN, ENN Algorithms on Eleven Different Datasets from UCI Machine Learning Repository

Mohammad Mahmudur Rahman Khan[1*], Rezoana Bente Arif[2#], Md. Abu Bakr Siddique[2@], and Mahjabin Rahman Oishe[3$]

[1]Dept. of ECE, Mississippi State University, Mississippi State, MS 39762, USA
[2]Dept. of EEE, International University of Business Agriculture and Technology, Dhaka 1230, Bangladesh
[3]Dept. of CSE, Rajshahi University of Engineering and Technology, Rajshahi 6204, Bangladesh
mrk303@msstate.edu[*], rezoana@iubat.edu[#], absiddique@iubat.edu[@], mahjabinoishe@gmail.com[$]

*Abstract*—Machine learning qualifies computers to assimilate with data, without being solely programmed [1, 2]. Machine learning can be classified as supervised and unsupervised learning. In supervised learning, computers learn an objective that portrays an input to an output hinged on training input-output pairs [3]. Most efficient and widely used supervised learning algorithms are K-Nearest Neighbors (KNN), Support Vector Machine (SVM), Large Margin Nearest Neighbor (LMNN), and Extended Nearest Neighbor (ENN). The main contribution of this paper is to implement these elegant learning algorithms on eleven different datasets from the UCI machine learning repository to observe the variation of accuracies for each of the algorithms on all datasets. Analyzing the accuracy of the algorithms will give us a brief idea about the relationship of the machine learning algorithms and the data dimensionality. All the algorithms are developed in Matlab. Upon such accuracy observation, the comparison can be built among KNN, SVM, LMNN, and ENN regarding their performances on each dataset.

*Keywords—Pattern recognition, Convex optimization, Semidefinite programming, Mahalanobis distance, Metric learning, Multi-class classification, Supervised learning, Intraclass coherence, Nearest neighbors, Maximum gain.*

## I. INTRODUCTION

In machine learning, supervised learning is the endeavor of learning an outcome from a labeled training dataset comprising a set of learning instances [4] so that machines can predict the proper output of test datasets. Supervised learning algorithms can be used for classification as well as regression. Most commonly used supervised learning algorithms are neural networks [5], k-nearest neighbors [6], support vector machine [7], naïve Bayes classifier [3], large margin nearest neighbor [8], and extended nearest neighbor [9].

It is beyond dispute that the machine learning algorithms demonstrates varying performance level for different datasets. A generalized guideline for applying these algorithms is yet to be established. Therefore, the primary motivation of this work is to observe the relationship among the datasets' features and the machine learning algorithms' performance.

In pattern recognition, k-nearest neighbors (KNN) [6] algorithm is the earliest and most straightforward nonparametric method based on nearest neighbor approach [10] used for classification and regression [8, 11]. In KNN classification, an unlabeled object is categorized by the majority labeled objects within its k adjacent neighbors in the training set. Though the significant values of k lessen the effect of noise on classification [12], it makes border between groups less clear-cut. In binary classification, the best value of k is selected empirically via bootstrap method [13].

Extended nearest neighbor (ENN) is the most recent supervised learning algorithm for pattern recognition that predicts the pattern of an unknown test sample hinged on the highest gain of intraclass coherence [9]. Unlike the KNN based classification method where only nearby neighbors are considered, ENN exploits the information from all available data to make a classification decision. ENN consider not only who the adjacent neighbors of the trial sample are, but also who envisage trial sample as their adjacent neighbors [9]. ENN, therefore, learns from the global distribution of data and thus improves classification performance. In addition, the modified versions of ENN, such as principal component analysis (PCA), binary particle swarm optimization (BPSO) and evolutionary algorithm (EA) based ENN (BPSO-EA-ENN) methods provide more accuracies than the pedigreed ENN [14].

Support vector machine (SVM) is a supervised learning model designed to achieve high performances in practical applications that makes classification by detecting the perfect hyperplane that boosts the margin of separation within the two categories. Though SVM constructs complex model and algorithms (contains a wide-ranging class of neural nets, radial basis function nets, and polynomial classifiers, etc. [15]); it is simple to analyze mathematically, because it maps the inputs into a high dimensional feature space through some nonlinear mapping in the input space. Even though it builds a linear algorithm in the high dimensional feature space, it does not make any computations in that high dimensional space. All necessary calculations are performed directly in the input space through some kernel tricks [15]. Because of SVM being the binary classifier, it is applicable to recognize the loads in the power system [16]. Again, Hybrid Multiclass SVM (HM-SVM) is used to uplift the quality of the identification of image [17].

Large margin nearest neighbor (LMNN) is a statistical machine learning algorithm that learns Mahanalobis distance in a supervised method to enhance the classification accuracy of KNN [8]. The algorithm is built on semidefinite programming, a subfield of convex optimization. In the

LMNN method, the metric is learned with the objective that k-nearest neighbors must correspond to the identical class whereas a large margin separates examples from different classes. Mahanalobis distances are calculated by linear transfiguration of the input space and then computing Euclidean distances in the transfigured space. The Euclidean distances in the transfigured spaced can equivalently be considered as Mahanalobis distances in the original space. The linear transformation of input space is derived by minimizing a loss function that made up of two terms. The first term reduces vast distances between examples in the matching class that is expected as k-nearest neighbors, whereas the second term increases small distances between samples with different courses. Adjusting these terms leads to a linear transfiguration of the input space that increase the amount of training samples whose k-nearest neighbors have matching classes. LMNN algorithm has many parallels to support vector machines (SVM), such as the metric learning minimizes to a convex optimization problem established on the hinge loss. But unlike SVM, LMNN does not require modification for difficulties in multi-way classification.

This paper has developed Matlab modeling of KNN, SVM, LMNN and ENN algorithms. These algorithms are trained on several training datasets. After that their performances are evaluated on test sets. In this paper, 11 datasets are used namely segmentation, seeds, Pima Indians diabetes, page blocks, Parkinson, movement libras, mammographic masses, knowledge, ionosphere, glass, and CNAE9. All the datasets are available in the UCI machine learning repository [18-20]. Finally, the variations of accuracies of these four algorithms were observed for every the dataset to make a comparison in the algorithms' performances.

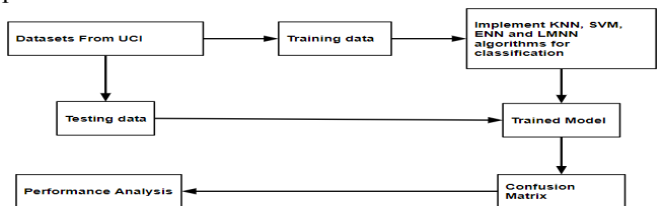

Fig. 1. The block diagram representing the steps done in this paper for examining the performances of the algorithms on various datasets.

## II. LITERATURE REVIEW

KNN and SVM are two fundamental algorithms to classify different prototypes in neural network whereas ENN and LMNN are quit new processes to perform these kinds of classifications. The uplift of the neural network in the bio-medical sector is becoming a benediction for us [21]. KNN and SVM algorithms are resembled as in both cases in a specific area which is considered to find the majority of probabilities. The improved KNN shows better accuracy on the location of the web text in clusters than automatic KNN algorithm [22]. Besides, KNN can be employed in criminological investigations to identify the glass type. In this case, the highest accuracy was observed for Boosting method which is 75.6% [23]. But in our paper, we found the highest accuracy 90.48% for 'Seeds' dataset in KNN. Though KNN is popular for its simplicity and effectiveness, few are lacking regarding some features like memory consumption as well as performance time. Thus, in many types of research, different methods using various datasets to develop its features have been proposed [24]. Another algorithm to classify data is 'Support Vector Machine' (SVM). The primary benefit of SVM is the capability of dealing with large number datasets which KNN fails to perform in some cases. On the other hand, the drawback of it is the high price and complexity [25]. In one previous research, KNN and SVM performances in water quality detection were observed using 10-fold Cross-Validation. The average accuracy for SVM and KNN was found 92.40% as well as 71.28% respectively [26]. However, in this paper, we have found 95.06% and 95.8% accuracy in SVM and KNN respectively. To overcome the inconveniences in KNN, another method named ENN was proposed, and its performance was observed which implied that ENN showed a better performance than the traditional KNN. In ENN, along with considering the nearest data test sample is also counted in the nearest data which gives a better accuracy [9]. This two-way communication style is further accurate than of KNN and SVM method. Another advanced method of the classical KNN is 'Large Margin Nearest Neighbor' (LMNN) which is exhibiting its powerful impact as a classifier in metrically related pattern recognition purposes. In various researches, the process of the implementation of LMNN classifier has been presented. On top of that, the error rates for this method were calculated which showed less error in its performance than KNN classifier [27, 28]. Though the LMNN algorithm is more complicated and expensive in real life application, its low error rate and better performance are replacing its drawbacks. In some research, some modifications in the LMNN method were proposed to minimize its complexity [29]. Besides, LMNN shows better accuracy than of LSML, LDML, LFDA, and LFDA in face reorganization [30]. In this paper, the accuracies for different datasets of LMNN were observed, and it was traced that LMNN gave excellent performance in its accuracy than the accuracies of other classifiers in every dataset furthermore. As the LMNN classifier provides a high range of efficiency in any datasets, it can be regarded the superior classifier among all of the four classifiers mentioned in this paper if its (LMNN) high price and compound practical implementation can be overcome.

## III. OVERVIEW OF ALGORITHMS

### A. The Basics of K-Nearest Neighbors (KNN)[6]

K-nearest neighbor (KNN) is a nonparametric and instance based lazy learning algorithm used to prognosticate the hierarchy of a new training value in a dataset where data values are segregated into a handful of classes. As KNN stores all the available cases and requires checking whole data set to classify a new sample point, the minimal training but extensive testing phase of KNN comes both at the memory and computational costs. KNN is a supervised training consisting of a given labeled dataset containing training observations (x, y) and would like to represent the correlation between x and y. The goal of KNN is to capture a function $h: x \rightarrow y$ so that given a new training value x, h(x) can confidently determine the corresponding output y. In KNN classification, a new training point is categorized by the higher number of votes of its neighbors, with the sample point being allocated to the group most common among its k nearest neighbors. If k=1, then new point is allocated to the group of its single nearest neighbor. Figure 2 shows the basic illustration of KNN classification.

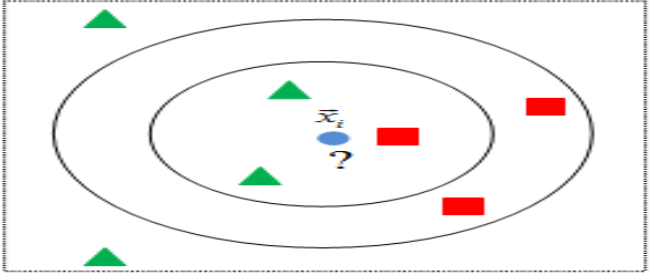

Fig. 2. The basic illustration of KNN classification. If k = 3 the test sample (blue circle) is allocated to the group of green triangle and if k = 5 then it is allocated to the group of red quadrilateral

The performance of KNN algorithm mainly relies on the distance metric employed to pinpoint the k adjacent neighbors of a sample point. The most frequently used one is the Euclidean distance given by:

$$d(x,x') = \sqrt{(x_1 - x'_1)^2 + (x_2 - x'_2)^2 + ..... + (x_n - x'_n)^2} \quad ...... (1)$$

For a given number of nearest neighbors, k, and an unknown sample point, x, and a distance metric d, a KNN classifier performs the following two functions:
1. It searches through the entire dataset and computes d between x and each training observation. Let assume that the k points in the training data that are adjacent to x belong to set U. It is noteworthy that k usually to be an odd number as it prevents the tied situation.
2. Then it calculates the conditional probability for each class, i.e., the possibility for a fraction of points present in the set U for a given class label. Finally, the unknown sample point x is allocated to the group with the highest probability.

$$P(y = j | X = x) = \frac{1}{K} \sum_{i \in U} I(y^{(i)} = j) \quad ...... (2)$$

B. *The Basics of Extended Nearest Neighbors (ENN)*

ENN classifies a new training value based on the highest gain of intra-class coherence. Unlike KNN, in which exclusively the adjacent neighbors of a test point are considered for classification, ENN classifies a new sample value by not only considering who are the adjoining neighbors of the test value, but also who consider the test values as their adjoining neighbors [9]. As KNN approach is sensitized to the variation of the distributions of the predefined classes, two kinds of error may occur, such as for the samples in the areas of higher density, k nearest neighbors actually may lie in the areas of slightly lower frequency or vice versa [31]. The bordering neighbors of an unknown observation tend to be dominated by the classes with higher density, which may bring about misclassification in KNN [31]. To solve this deficiency in KNN, ENN exploits the generalized class wise statistics from all training data to learn from global data distribution and thus improves classification performance. Let the generalized class wise statistic $T_i$ for class *i* regarding the pooled samples $S_1$ and $S_2$ for each category along with its nearest neighbors can be expressed as:

$$T_i = \frac{1}{n_i k} \sum_{x \in S_i} \sum_{r=1}^{k} I_r(x, S = S_1 \cup S_2) \; ; i = 1, 2 \quad ...... (3)$$

Where, $S_1$ and $S_2$ = The samples in classes 1 and 2 respectively

$x$ = One single sample in $S$
$n_i$ = Number of samples in $S_i$
$k$ = The number of adjacent neighbors
$I_r(x,S)$ = Indicator function

The indicator function $I_r(x,S)$ determines whether both the sample x and its rth nearest neighbors pertain to the matching category or not, can be expressed as:

$$I_r(x,S) = \begin{cases} 1, & \text{if } x \in S_i \text{ and } NN_r(x,S) \in S_i \\ 0, & \text{otherwise} \end{cases} \quad ...... (4)$$

Where $S_i$ = The samples in class i
$NN_r(x, S)$ = The rth nearest neighbor of x in S

Note that the generalized class wise statistic $T_i$ in equation (1) measures the coherence of data from the same class and $0 \leq T_i \leq 1$ with $T_i = 1$ when all the bordering neighbors of class I are from the same class i and $T_i = 0$ when all the nearest neighbors are from different classes.
As $T_i$ represents the data distribution across multiple classes, the intraclass coherence can be demonstrated as follows:

$$\Theta^j = \sum_{i=1}^{2} T_i^j \quad ...... (5)$$

Now to classify an unknown sample Z, it is iteratively assigned to classes 1 and 2 respectively to obtain two new generalized class wise statistics $T_i^j$, where j = 1,2.

$$T_i^j = \frac{1}{n'_i k} \sum_{x \in S'_{i,j}} \sum_{r=1}^{k} I_r\left(x, S' = S_1 \cup S_2 \cup \{Z\}\right) \; ; i, j = 1, 2 \quad ...... (6)$$

Where, $n'_i$ = The size of $S'_{i,j}$

$$S'_{i,j} = \begin{cases} S_i \cup \{Z\}, & \text{when } j = i \\ S_i, & \text{when } j \neq i \end{cases}$$

Then the ENN categorizes the sample Z according to the following target function:

$$f_{ENN} = \arg\max_{j \in 1,2} \sum_{i=1}^{2} T_i^j = \arg\max_{j \in 1,2} \Theta^j \quad ...... (7)$$

Now, the two class ENN classification method can be easily extended to multi-class classification by:

$$f_{ENN} = \arg\max_{j \in 1,2,...,N} \sum_{i=1}^{N} T_i^j \quad ...... (8)$$

To avoid the recalculation of generalized class wise statistics $T_i^j$ in the testing stage, the sample Z is iteratively assigned to each possible class j, j=1, 2, …., N, and envisage the class association conforming to an equivalent target function of equation (6), defined as:

$$f_{ENN.V} = \arg\max_{j \in 1,2,...,N} \left\{ \left(\frac{\Delta n_i^j + k_i - k T_i}{(n_i + 1)k}\right)_{i=j} - \sum_{i \neq j}^{N} \frac{\Delta n_i^j}{n_i k} \right\} \quad ...... (9)$$

Where,
K = The custom parameter of the number of the adjacent neighbors defined by the user

$n_i$ = the numbers of training data for class i
$K_i$ = The number of the bordering neighbors of the test sample Z from class i
$\Delta n_i^j$ = the change of the k nearest neighbors for class i when the test sample Z is assumed to be class j
$T_i$ = The generalized class wise statistic of original class i (i.e., test sample Z is not included).

### C. The Basics of Support Vector Machine (SVM)

In a supervised classification problem, data need to be separated into training and testing sets. Each sample point in the training set consists of a class label and several features or observed variables. The objective of SVM is to build a model based on training data samples to prognosticate the class labels of test data given solely test data features. SVM algorithm performs linear categorization by obtaining the hyperplane that boosts the margin between two classes. The data points that determine the hyperplane are support vectors. In essence, for a set of given labeled training data, the SVM algorithm obtains a most favorable hyperplane which classifies new test data. Figure 3 shows the Basic illustration of SVM classification.

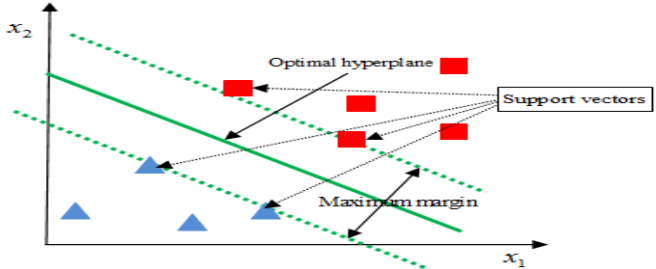

Fig. 3. The basic illustration of SVM classification showing the maximum margin, optimal hyperplane, and the support vectors

For a given training dataset of n points of the form $(\vec{x}_1, y_1), \ldots, (\vec{x}_n, y_n)$ where $y_i = \pm 1$ indicating the class belongingness of the set of points $\vec{x}_i$. Now, a hyperplane can be demonstrated by the number of values $\vec{x}_i$ as:

$$\vec{w} \cdot \vec{x}_i + b = 0 \quad \ldots\ldots (10)$$

To get the optimal hyperplane that separates the set of values $\vec{x}_i$ into either $y_i = +1 \text{ or } -1$, the separation between the hyperplane and the closest point $\vec{x}_i$ from either group should be maximized. So, the maximum margin hyperplane can be defined as:

$$\vec{w} \cdot \vec{x}_i + b = \pm 1 \quad \ldots\ldots (11)$$

Now the distance of a data point and hyperplane can be written as:

$$d = \frac{|\vec{w} \cdot \vec{x}_i + b|}{\|\vec{w}\|} \quad \ldots\ldots (12)$$

Again, by putting the numerator equal to one from equation (11) in equation (12), the gap to the support vectors from a hyperplane is given by:

$$d_{\text{support vectors}} = \frac{1}{\|\vec{w}\|} \quad \ldots\ldots (13)$$

As hyperplane is defined by a two-class problem for $y_i = \pm 1$, the margin M is twofold the distance to the nearest examples:

$$\therefore M = \frac{2}{\|\vec{w}\|} \quad \ldots\ldots (14)$$

Finally, the issue of optimizing M is parallel to the issue of minimizing $\|\vec{w}\|$. To solve this problem, the SVM algorithm [7, 32] necessitates the resolution of the following optimization problem subject to some constraints:

$$\min_{w,b,\xi} \frac{1}{2} w \cdot w + C \sum_{i=1}^{n} \xi_i \quad \ldots\ldots (15)$$

subject to constraints: (i) $y_i (w \cdot \varnothing(x_i) + b) \geq 1 - \xi_i$
(ii) $\xi_i \geq 0$

Here the function $\varnothing$ represents the mapping of training data $\vec{x}_i$ into a higher dimensional linearly separable space if the training data $\vec{x}_i$ are nonlinear in lower dimensional space. Then SVM algorithm obtains a linear segregating hyperplane in the more upper dimensional space. C is the penalty parameter of the error function which is habitually higher than zero.

The restricted optimization problem of equation (15) can be converted to an unrestricted optimization problem by using the Lagrangian function as given by equation 16 [7].

$$L(w,b,\xi) = \frac{1}{2} w \cdot w + C \sum_{i=1}^{n} \xi_i - \sum_{i=1}^{n} \alpha_i [y_i(w \cdot \varnothing(x_i) + b) - 1 + \xi_i] - \sum_{i=1}^{n} r_i \xi_i \quad \ldots (16)$$

After optimization of equation (16), the classification decision of a new test data Z can be determined by observing the sign of equation (17). If the sign is positive, the sample Z belongs to class 1 and if the sign is negative Z belongs to class 2.

$$D(\vec{Z}) = \text{sgn}\left(\sum_{j=1}^{n} \alpha_j y_j \varnothing(\vec{x}_j) \cdot \vec{Z} + b\right) \quad \ldots\ldots (17)$$

Again, $K(x_i, x_j) \equiv \varnothing(x_i) \cdot \varnothing(x_j)$ is called the kernel function.

### D. The Basics of Large Margin Nearest Neighbors (LMNN)

LMNN is a machine learning algorithm that gets the hang of a Mahalanobis distance that maximizes KNN classification performance. The objective of LMNN is to learn an optimized metric (Mahalanobis metric) such that k-adjoining neighbors always pertain to the same category whereas a large margin segregates examples from different categories. Figure 4 shows the basic illustration of LMNN classification.

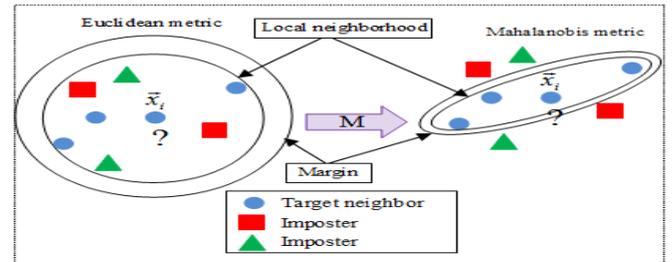

Fig. 4. The basic illustration of LMNN classification where adapting the Mahalanobis distance metric instead of the Euclidean metric increases the performance of the classifier

Let the training data set of n tagged examples be:
$$S = \{(\vec{x}_i, y_i)\}_{i=1}^{n} \subset R^d \quad \ldots\ldots (18)$$
Where $y_i$ = Discrete class levels

The goal of the algorithm is to study a linear transformation L: $R^d \to R^d$, which will be used to compute the square distances as:
$$D(\vec{x}_i, \vec{x}_j) = \|L(\vec{x}_i - \vec{x}_j)\|^2 \quad \ldots\ldots (19)$$

The algorithm distinguishes between two types of data samples:
1. <u>Target neighbors:</u> Each input $x_i$ has precisely k different target neighbors within S with the identical class label $y_i$. It is desired that all the k target neighbors have minimal distance to $x_i$, as computed by equation (19). All target neighbors for a given input data $x_i$ are supposed to be nearest neighbors under the learned metric. $\eta_{ij} \in \{0,1\}$ is used to determine whether sample $x_j$ is a target neighbor of input sample $x_i$.
2. <u>Imposters:</u> An imposter of an input sample $x_i$ is another sample $x_j$ with a different class label (i.e., $y_i = y_j$) which is one of the adjacent neighbors of $x_i$. During the learning period, the algorithm tries to minimize the number of imposters for all data input in the training set.

The cost function for LMNN over the distance metrics expressed in equation (19) has two competing terms. The initial term imposes a penalty on vast distances between each input and its target neighbors, while the second term imposes a penalty on small gaps between each input and all the imposters. The cost function is given by:
$$\varepsilon(L) = \sum_{ij} \eta_{ij} \|L(\vec{x}_i - \vec{x}_j)\|^2 + c\sum_{ijl} \eta_{ij}(1-y_{il})\left[1 + \|L(\vec{x}_i - \vec{x}_j)\|^2 - \|L(\vec{x}_i - \vec{x}_l)\|^2\right]_+ \quad \ldots\ldots (20)$$

In equation (20), the second term $[z]_+ = \max(z,0)$ represents the hinge loss and constant $C > 0$ which is usually set by cross-validation. For each input $x_i$, the hinge loss is incurred by differently labeled imposters who are within one absolute unit of distance, the distance from input $x_i$ to any of its target neighbors. The cost function thereby helps to find optimized distance metrics so that the differently labeled samples maintain a large margin of distance and do not invade each other's neighborhood.

To determine the global minimum of equation (20) efficiently, the optimization of equation (20) can be redeveloped as an example of semidefinite programming [8]. To get the equivalent SDP, the equation (19) can be rewritten as:
$$D(\vec{x}_i, \vec{x}_j) = (\vec{x}_i - \vec{x}_j)^T M (\vec{x}_i - \vec{x}_j) \quad \ldots\ldots (21)$$

Where the matrix $M = L^T L$ = The Mahanalobis distance metric

By introducing slack variables $\xi_{ij}$ for all pairs of differently labeled inputs (i.e., $i \neq j$), the equation (20) can be written as an SDP regarding M as following as equation (22) [8].

$$\min \sum_{ij} \eta_{ij} (\vec{x}_i - \vec{x}_j)^T M(\vec{x}_i - \vec{x}_j) + c\sum_{ij} \eta_{ij}(1-y_{il})\xi_{ijl} \quad \ldots\ldots (22)$$

subject to constraints: (1) $(\vec{x}_i - \vec{x}_l)^T M(\vec{x}_i - \vec{x}_l) - (\vec{x}_i - \vec{x}_j)^T M(\vec{x}_i - \vec{x}_j) \geq 1 - \xi_{ijl}$
(2) $\xi_{ijl} \geq 0$
(3) $M \geq 0$

Here, the slack variables $\xi_{ij}$ minimize the number of violations of the imposter. The 3rd constraint ensures that M is a positive semidefinite.

IV. RESULTS AND DISCUSSION

The accuracies of KNN, ENN, SVM, and LMNN in different datasets are represented in the column graph as shown in figure 4 as well as in table 1. From the graph, it can be illustrated which dataset gives the better accuracy.

At first, it is significantly shown that among all the datasets, considered for different classifiers, LMNN gives the highest accuracy for the 'Knowledge' dataset. On the contrary, the lowest accuracy was found for 'Glass' dataset in ENN which is 0.6744. It is also clearly shown that in any dataset the pick accuracy was observed in LMNN.

Another important point is that both maximum and minimum accuracies were found in KNN and ENN for 'Page Blocks' and 'Glass' dataset respectively. Again, the lowest efficiency for these two classifiers is the same which is 0.6744.

Coming to the performance of the other two classifiers, SVM and LMNN, the maximum accuracies in the performance of SVM and LMNN were found in 'CNAE9' and 'Knowledge' dataset respectively. Though the lowest accuracies in KNN, ENN and SVM are around 0.6, in the LMNN the lowest accuracy is 0.9929 which is higher than the peak accuracy among the rest of the three classifiers.

The LMNN shows the higher accuracy in any dataset than of the KNN, ENN, and SVM classifier.

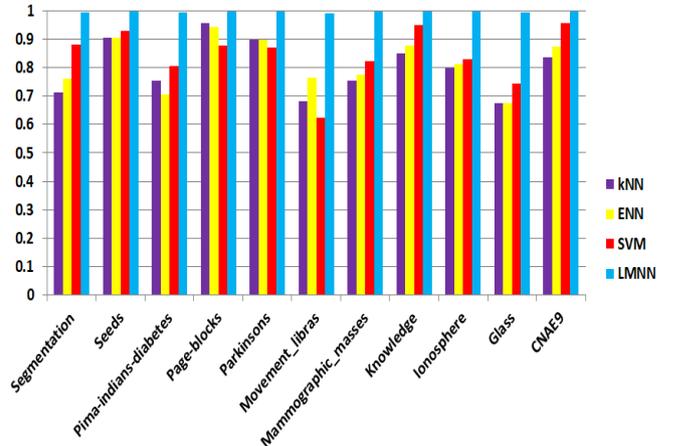

Fig. 5. Accuracy levels of KNN, ENN, SVM, and LMNN algorithms for eleven datasets from the UCI machine learning repository

Now, the table 1 below bears the evidence that the performance of the algorithms varies for different datasets due to their feature dimensions and range. However, the overall performance of LMNN algorithm is better compared to other algorithms which mean that the distance matrix measurement has a significant impact upon the classification accuracy. Contrarily, considering the computational complexity, the ENN and KNN are less complicated compared to the SVM and LMNN. The LMNN has a considerable computational cost due to the calculation of the distance matrix.

TABLE I. PERFORMANCES OF DIFFERENT ALGORITHMS FOR VARIOUS DATASETS

| Datasets | KNN | ENN | SVM | LMNN |
|---|---|---|---|---|
| Segmentation | 0.7143 | 0.7619 | 0.881 | 0.9967 |
| Seeds | 0.9048 | 0.9048 | 0.9286 | 0.9993 |
| Pima-Indians-diabetes | 0.7532 | 0.7078 | 0.8052 | 0.9963 |
| Page-blocks | 0.958 | 0.9443 | 0.8776 | 0.9992 |
| Parkinsons | 0.8974 | 0.8974 | 0.8718 | 0.9987 |
| Movement_libras | 0.6806 | 0.7639 | 0.625 | 0.9929 |
| Mammographic_masses | 0.7552 | 0.776 | 0.8229 | 0.9977 |
| Knowledge | 0.8519 | 0.8765 | 0.9506 | 0.9996 |
| Ionosphere | 0.8 | 0.8143 | 0.8286 | 0.9971 |
| Glass | 0.6744 | 0.6744 | 0.7442 | 0.9956 |
| CNAE9 | 0.838 | 0.875 | 0.9583 | 0.9983 |

## V. CONCLUSION

All these four algorithms in machine learning have been showing their powerful impacts on the classification of data in various sectors. Though currently KNN and SVM are replaced by ENN and LMNN, due to more perfection and accuracy, the vital role of KNN and SVM cannot be ignored. In this paper, the accuracies of these four classifiers for twelve datasets were observed using MNIST datasets. All of these four classifiers gave the accuracy of more than 90% for different datasets. Moreover, LMNN classifier showed accuracies more than 99% for all datasets. Therefore, even though the simplicity and smooth implementation of the rest of the three classifiers, LMNN is more accurate to be selected due to its higher accuracies. LMNN showed the best performance (99.96% accuracy) for 'knowledge' data-set. Thus, we are planning to improve this accuracy in LMNN with a less costly method for 'knowledge' data-set in our future research.